\documentclass[letterpaper, 10 pt, conference]{ieeeconf}  

\IEEEoverridecommandlockouts                              

\usepackage{graphics} 
\usepackage{epsfig} 
\usepackage{mathptmx} 
\usepackage{times} 
\usepackage{amsmath} 
\usepackage{amssymb}  
\usepackage{xcolor}
\usepackage{graphicx}
\usepackage{subcaption}
\usepackage{ifthen}
\usepackage{soul}
\usepackage{xcolor}
\usepackage[linesnumbered,ruled,vlined]{algorithm2e}
\usepackage{multirow, makecell}
\let\labelindent\relax
\usepackage{enumitem}
\usepackage{breqn}
\usepackage{setspace}
\usepackage{booktabs}
\usepackage{microtype}
\usepackage{float}
\usepackage[utf8]{inputenc}

\newboolean{showremoved}

\let\oldst\st 
\renewcommand{\st}[1]{%
  \ifthenelse{\boolean{showremoved}}%
    {\oldst{#1}}
    {}
}

\title{\LARGE \bf
AeroSafe: Mobile Indoor Air Purification using Aerosol Residence Time Analysis and Robotic Cough Emulator Testbed
}

\author{M Tanjid Hasan Tonmoy$^{1}$, Rahath Malladi$^{2}$, Kaustubh Singh$^{2}$, Forsad Al Hossain$^{3}$,\\ Rajesh Gupta$^{1}$, Andrés E. Tejada-Martínez$^{4}$ and Tauhidur Rahman$^{1}$%
\thanks{$^{1}$ University of California San Diego, La Jolla, CA 92093, USA}%
\thanks{$^{2}$ Plaksha University, Punjab 140306, India}%
\thanks{$^{3}$ University of Massachusetts Amherst, Amherst, MA 01003, USA}%
\thanks{$^{4}$ University of South Florida, Tampa, FL 33620, USA}%
\thanks{© 2025 IEEE. This is the author’s version of the paper. It is posted here for personal use. Not for redistribution. The definitive version will appear in the Proceedings of the IEEE International Conference on Robotics and Automation (ICRA), 2025.}
}

\begin{document}

\maketitle
\thispagestyle{empty}
\pagestyle{empty}

\begin{abstract}

Indoor air quality plays an essential role in the safety and well-being of occupants, especially in the context of airborne diseases. This paper introduces AeroSafe, a novel approach aimed at enhancing the efficacy of indoor air purification systems through a robotic cough emulator testbed and a digital-twins-based aerosol residence time analysis. Current portable air filters often overlook the concentrations of respiratory aerosols generated by coughs, posing a risk, particularly in high-exposure environments like healthcare facilities and public spaces. To address this gap, we present a robotic dual-agent physical emulator comprising a maneuverable mannequin simulating cough events and a portable air purifier autonomously responding to aerosols. The generated data from this emulator trains a digital twins model, combining a physics-based compartment model with a machine learning approach, using Long Short-Term Memory (LSTM) networks and graph convolution layers. Experimental results demonstrate the model's ability to predict aerosol concentration dynamics with a mean residence time prediction error within 35 seconds. The proposed system's real-time intervention strategies outperform static air filter placement, showcasing its potential in mitigating airborne pathogen risks.

\end{abstract}


\section{INTRODUCTION}

The concentration of various airborne particles heavily influences the safety and comfort of individuals in indoor spaces. Existing operational strategies for indoor air purifiers fail to consider the dynamic variations in particle concentration resulting from human respiratory events, such as coughs. However, these syndromic events impact the health and safety of occupants and should therefore be considered to improve the safety of people susceptible to exposure to airborne diseases such as COVID-19 or influenza. To address this dynamic challenge, integrating robotics offers a novel approach to real-time intervention. Autonomous robotic systems provide the precision, adaptability, and responsiveness required to mitigate the risk of airborne disease transmission, strategically reducing the time the particles generated by syndromic events linger in the air (referred to as residence time). Reducing airborne disease spread is crucial in high-risk spaces such as classrooms, auditoriums, and healthcare facilities~\cite{nih_airborne3, nih_airborne4}. Standard guidelines that overlook aerosol dispersion patterns may inadvertently facilitate the spread of infections. Therefore, considering the critical significance of indoor air quality \cite{dominguez-amarillo_bad_2020, nor2021particulate}, deployment of advanced indoor air-filtration systems is vital to mitigate associated health risks. Such systems should be capable of detecting aerosol-generating events such as coughing, forecasting the subsequent concentration changes, and initiating adaptive interventions to minimize the aerosol's residence time in the space. Robotic air purifiers, equipped with sensors and mobility, enable real-time, targeted air filtration, enhancing the system's efficiency in responding to localized aerosol threats.

\begin{figure*}[!ht]
\vspace{+1.5mm}
    \centering
    \includegraphics[width=\linewidth]{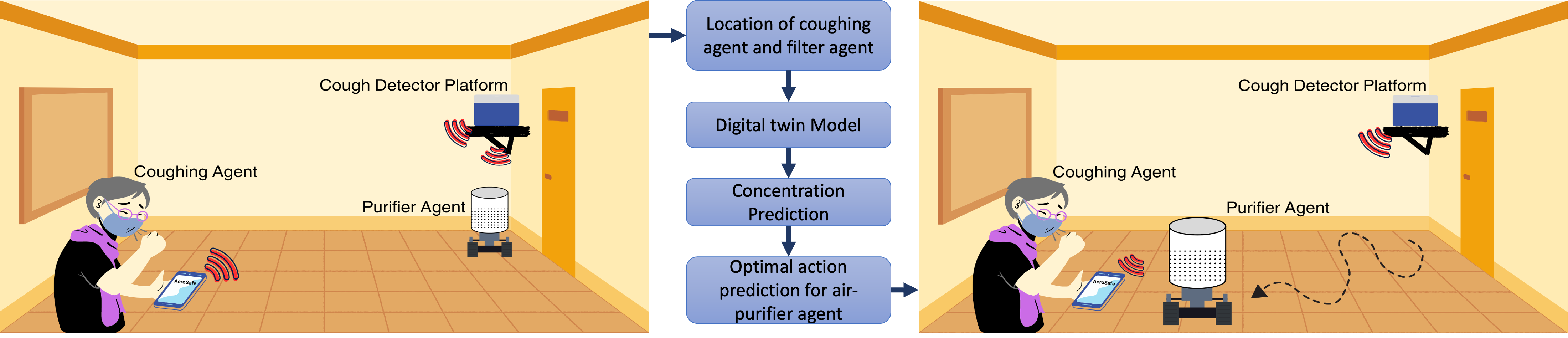}
    \caption{\textbf{Overview of AeroSafe system:} Real-time intervention by a mobile air purifier after detection of a cough event.}
    \vspace{-1em}
    \label{fig:overview}
\end{figure*}

A major challenge in designing adaptive autonomous air purifiers is the lack of data to train models to forecast aerosol concentration and residence times, as CFD simulations require significant resources and expertise \cite{cfd_comp, cfd_aero}. We address this by designing a robotic dual-agent emulator, including a robotic mannequin simulating cough and a mobile air purifier for mitigation. This approach allows for experimental validation of model learning and generalization across diverse environments. Our system, equipped with a sensor-instrumented testbed, enables controlled experimentation with consistent and repeatable measurements. Parameters such as HVAC settings, sensor placement, and robotic agent position can be adjusted to test various scenarios. The robotic cough emulation mannequin provides a consistent approach to emulate coughs, balancing the complexity of the hardware and the fidelity of the emulation \cite{lindsley_dispersion_2012, lindsley_cough_2013, patel_assessing_2020}. The preference for a robotic system over manual emulation methods ensures precision and consistency. We used the data collected to train digital twin models of the aerosol concentration.

The proposed digital twin model integrates a physics-based compartment model with machine learning modules, including a Long Short Term Memory (LSTM) network and graph convolution layers. The system features a centralized sensing unit and particulate matter(PM) sensors for model training and calibration. The central unit passively detects cough events and estimates their origins. In response, the robotic air purifier autonomously navigates the environment, adjusting its position based on the model's predicted aerosol concentrations to determine optimal placement positions to minimize aerosol mean residence time (MRT). While a grid of PM sensors is used for initial training, our digital twin model requires minimal ongoing data for calibration, making it suitable for real-world deployment without extensive sensor infrastructure. The model accurately predicts aerosol concentration dynamics, enabling MRT estimation across various scenarios. Our best model achieved MRT prediction with a mean error under $35$ seconds. Meta-learning-based training allows the model to adapt to new environments, demonstrating its generalizability in zero and few-shot scenarios. Our approach identifies optimal air purifier placements, achieving faster aerosol removal in both single and multiple cough scenarios.

\section{RELATED WORKS}

Numerous studies have explored aerosol dispersion from respiratory activities, particularly in light of the COVID-19 pandemic, aiming to assess infection risks through various methods, including complex simulations and sensor-based measurements \cite{virbulis_numerical_2021}. While simulations often rely on standardized representations of indoor spaces and sensor-based research is typically conducted in controlled environments, translating these findings to human-occupied spaces remains a challenge.


CFD simulations have been widely used to study indoor air quality, focusing on transmission \cite{lohner_detailed_2020, oh_numerical_2022} and mitigation strategies in classrooms \cite{foster_estimating_2021, abuhegazy_numerical_2020}, hospitals \cite{ren_numerical_2021, bhattacharyya_novel_2020}, and aircraft cabins \cite{zee_computational_2021}. Machine learning models trained on CFD data to predict droplet transmission in indoor spaces \cite{tamaddon_jahromi_predicting_2022}, and approaches for optimization of ventilation guidelines for buses \cite{mesgarpour_predicting_2022}, creating simulations of digital twins of CFD ~\cite{molinaro_embedding_2021} and digital twins for physical phenomena, such as fire propagation \cite{zohdi_machine-learning_2020} have come to light. To address the complexity of CFD simulations, simplified methods like the two-compartment model \cite{bathula_survival_2021} were proposed. These models assume perfect mixing within compartments, providing predictions lacking precision in long-term trends \cite{ganser2017models}. 

Sensor-based approaches have also been explored extensively such as \cite{virbulis_numerical_2021} which combines sensor data and simulations for real-time infection risk prediction. CO\textsubscript{2} sensors, used as proxies for exhaled air, have been investigated \cite{rusch_internet_2022}, along with similar sensor-based systems \cite{telicko_monitoring_2021, peladarinos_early_2021} and particulate matter (PM) sensors have been validated in environments like public transport \cite{fatih_covid-19_2020} and hospitals \cite{glenn_aerosol_2022}.

Portable air-cleaning systems have been studied for their efficacy in reducing pollutants \cite{hammond_should_2021}, with effectiveness examined across homes, offices \cite{lu_real-world_2023, derk_efficacy_2023}, hospitals \cite{dellweg_use_2022}, and larger other spaces \cite{zhai_application_2021}. The impact of purifier placement has been noted \cite{novoselac_impact_2009}, and alternative disinfection methods like ultraviolet light robots have been explored \cite{ultraviolet_2021}.

\begin{figure}[!b]
    \centering
    \begin{subfigure}{0.85\linewidth}
        \centering
        \includegraphics[width=\linewidth, height=3.2cm]{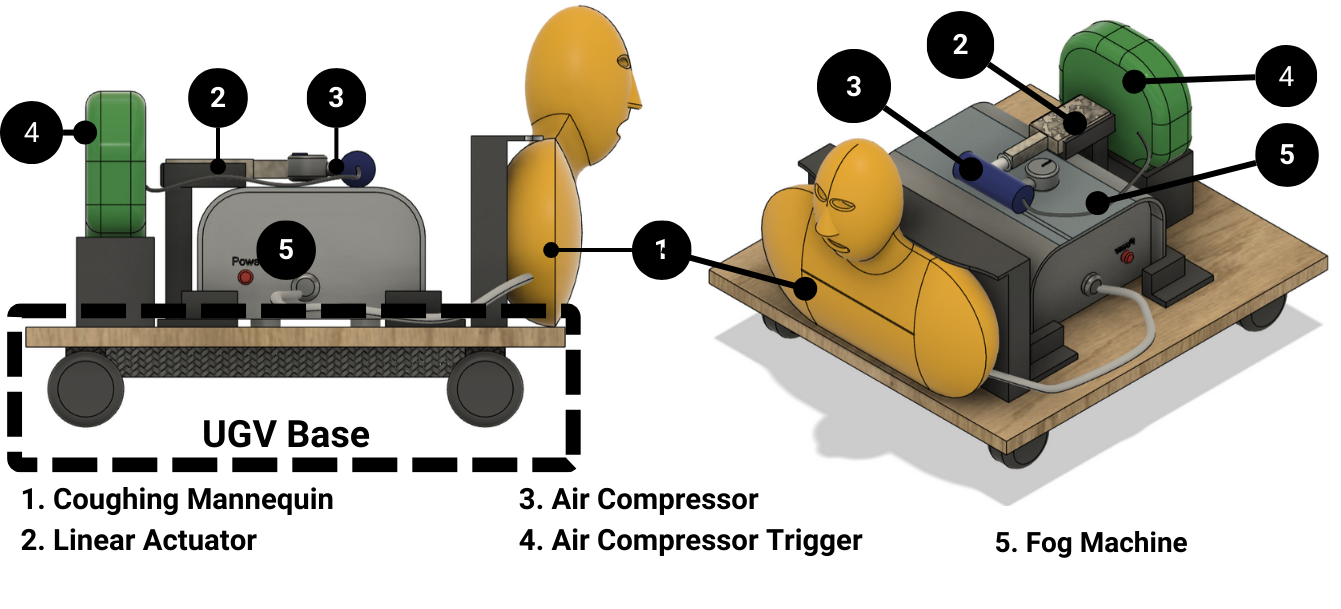}
        \caption{}
        \label{fig:cough_agent}
    \end{subfigure}
    \begin{subfigure}{0.8\linewidth}
        \centering
        \includegraphics[width=\linewidth, height=3.2cm]{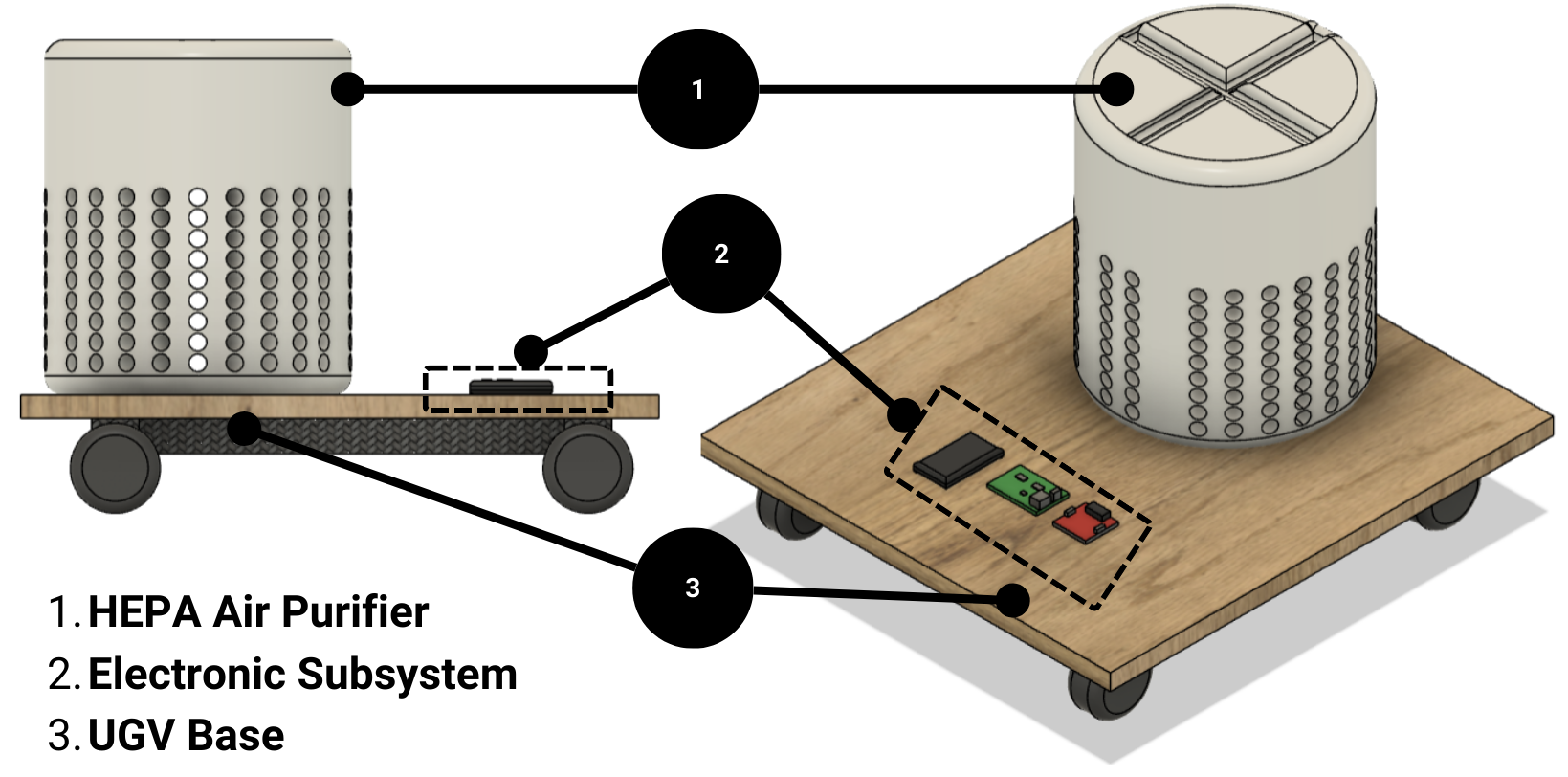}
        \caption{}
        \label{fig:filter_agent}
    \end{subfigure}
    \caption{CAD models of the two-agent system comprising (a) a coughing agent and (b) an air purifier agent.}
\end{figure}

\begin{itemize}
    \item \textbf{Novel Digital Twins System:} We propose a novel digital-twins-based solution that addresses the limitations of high-cost CFD simulations. Unlike previous works relying solely on CFD simulations for indoor air quality assessment \cite{lohner_detailed_2020}-\cite{zee_computational_2021}, our system integrates machine learning with physics-based models, validated using PM-sensor data, offering a more accessible and computationally efficient alternative.

    \item \textbf{Cough Emulation Testbed:} We introduce a two-agent robotic system where one robot simulates cough events, and the other deploys an air purifier to mitigate aerosol spread. This dynamic, real-time intervention system is a novel contribution.

    \item \textbf{Hybrid Physics-ML Model:} Our system enhances existing machine learning models integrated with simulation data \cite{tamaddon_jahromi_predicting_2022}- \cite{molinaro_embedding_2021}. By combining physics-based modeling with machine learning, our hybrid models improve predictive accuracy while adhering to the physical constraints of aerosol dispersion.

    \item \textbf{Optimized Air Purifier Placement:} Our research builds on prior investigations into portable air-cleaning systems \cite{hammond_should_2021}-\cite{ultraviolet_2021} by introducing a mobile air purifier robot, actively guided to optimize its placement for maximal efficiency in aerosol mitigation.
\end{itemize}

\begin{figure*}[!h]
\vspace{+1.5mm}
    \centering
    \begin{subfigure}{0.42\linewidth}
        \includegraphics[width=\linewidth, height=4.5cm]{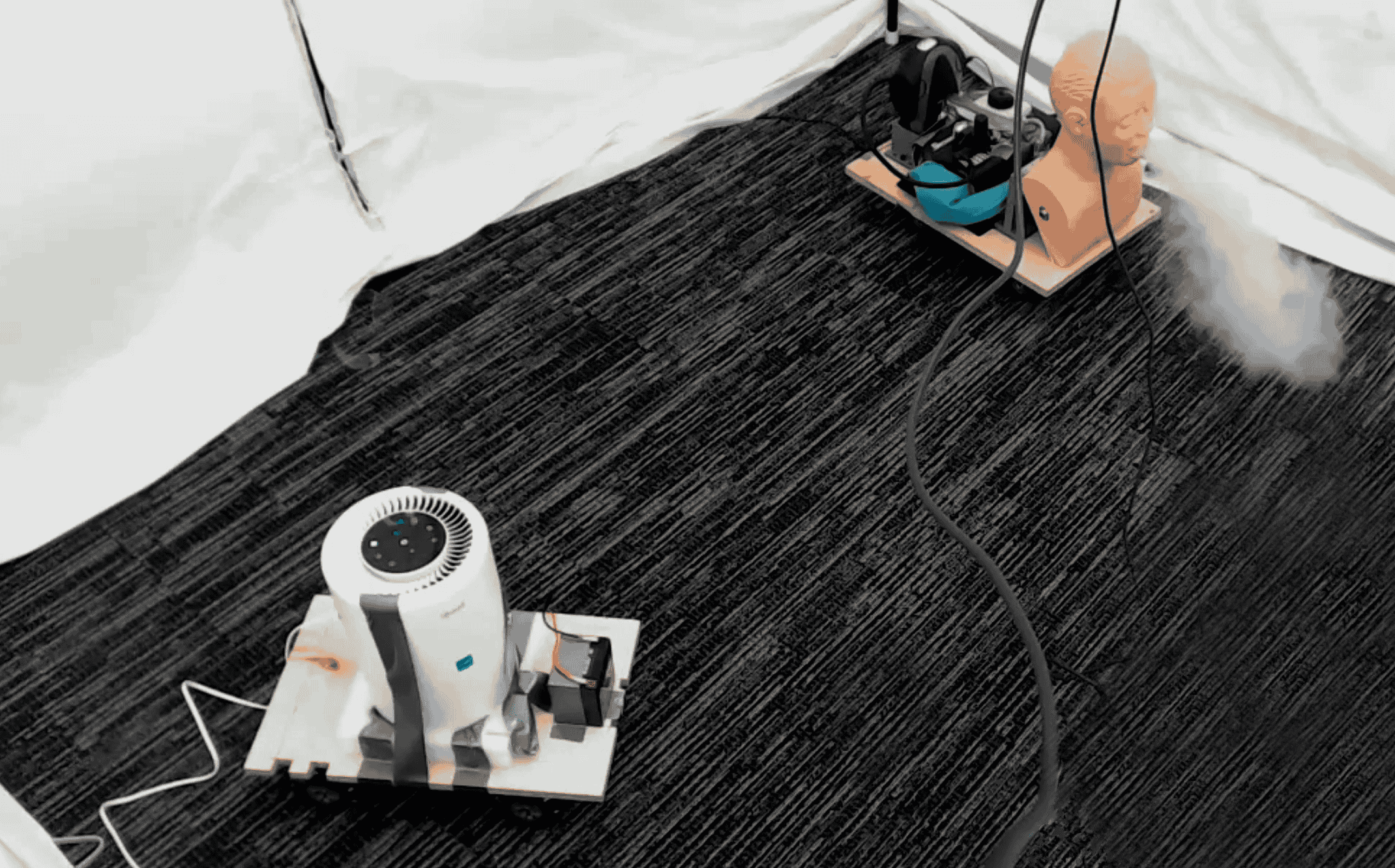}
        \caption{}
        \label{fig:Exp_Setup}
    \end{subfigure}
    \begin{subfigure}{0.45\linewidth}
    \centering
    \includegraphics[width=\linewidth, height=4.5cm]{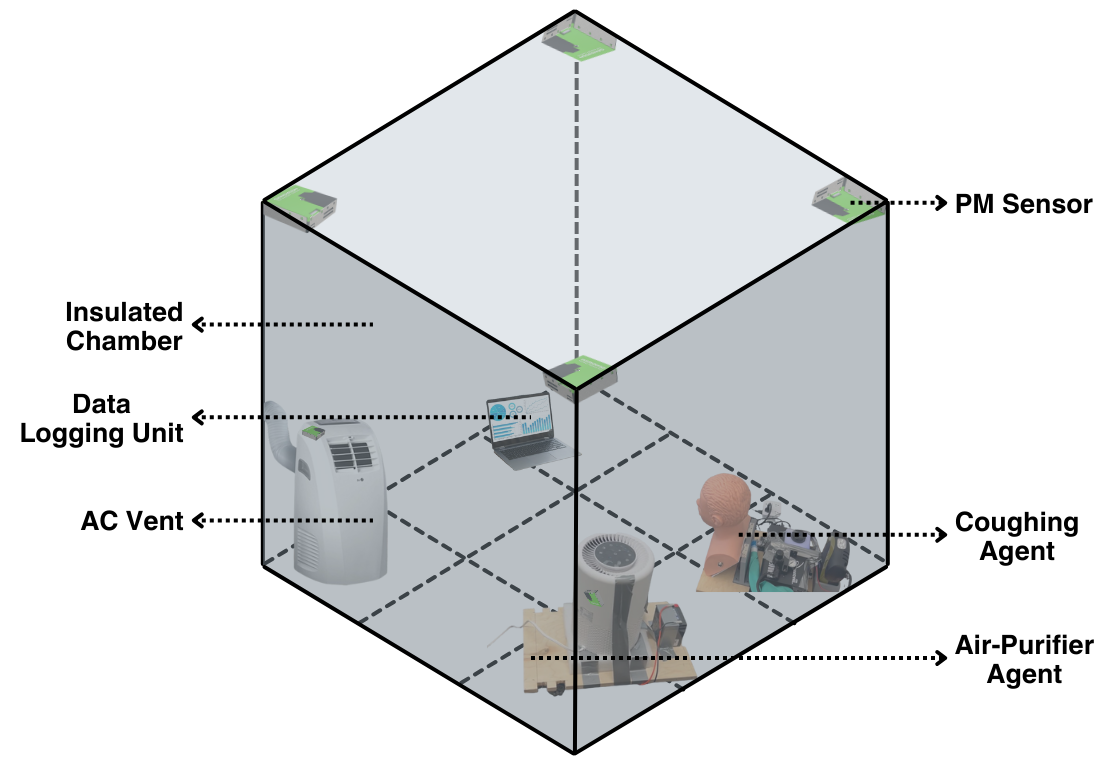}
    \caption{}
    \label{fig:two-agent-exp-setup}
    \end{subfigure}
    \caption{Experiment testbed setup - (a) Coughing and filter robot in testbed environment, (b) Schematic of testbed isolated with a plastic sheet showing grid discretization of the indoor space. (Image for representation purposes only.)}
    \vspace{-1em}
\end{figure*}

\section{EXPERIMENTAL SETUP}

\subsection{Sensors and Hardware}

\emph{\textbf{Dual-Agent Testbed:}} Our cough emulation system, based on \cite{zhou_development_2022}, utilizes a fog machine (Chauvet DJ Hurricane 1200) mounted on a robotic UGV for aerosol generation, an air compressor for cough-like emission, and a mechanical ventilator for controlled dispersion. Unlike the existing complex setups \cite{lindsley_dispersion_2012, lindsley_cough_2013}, our design is simplified (Fig.~\ref{fig:cough_agent}). The other robot, equipped with an air purifier (Levoit Smart True HEPA Core 200s), features adjustable fan speed via WiFi. Both agents are controlled by Raspberry Pis, with PM sensors placed throughout the testbed. A portable AC unit regulates ventilation and temperature in the isolated chamber, set up using plastic sheeting (shown in Fig.~\ref{fig:two-agent-exp-setup}).

\emph{\textbf{Sensors:}} We employ PM sensors (SPS30, Sensirion AG) strategically placed to measure aerosol concentrations. These sensors capture mass (µg/m³) and number concentrations (\#/cm³) for particles sized 1.0, 2.5, 4.0, and 10.0 microns. Additionally, a modified contactless sensing platform \cite{fluSense} with a 4-channel microphone array, Raspberry Pi, and Intel NCS 2 detects coughs and human presence.

\subsection{Experiment with Two Agent Testbed}
We designed a testbed as a $3\times3$ grid, representing possible locations for the robotic agent. This grid-based model enables generalization without exhaustive data collection from all continuous positions, which is both resource-intensive and offers minimal additional insights. Each grid cell simplifies the coughing agent’s orientation to cardinal directions (North, South, East, and West). Data was collected across varying configurations of the air conditioner (AC), air purifier, and coughing agent. Each data instance includes particle concentration (PM) readings for up to 15 minutes followed by an emulated cough. We systematically varied the agent locations and environmental conditions, including toggling the AC power and fan speeds. Initial trials were conducted with only the coughing agent in different positions, while the air purifier remained inactive. Subsequent trials introduced the air purifier robot in combination with the AC, with changes to furniture and testbed layout to generate diverse data for model training. 

\section{METHOD}

\begin{figure*}[!htbp]
\vspace{+1.5mm}
    \centering
    \includegraphics[width = .85\linewidth, height=1.4in]{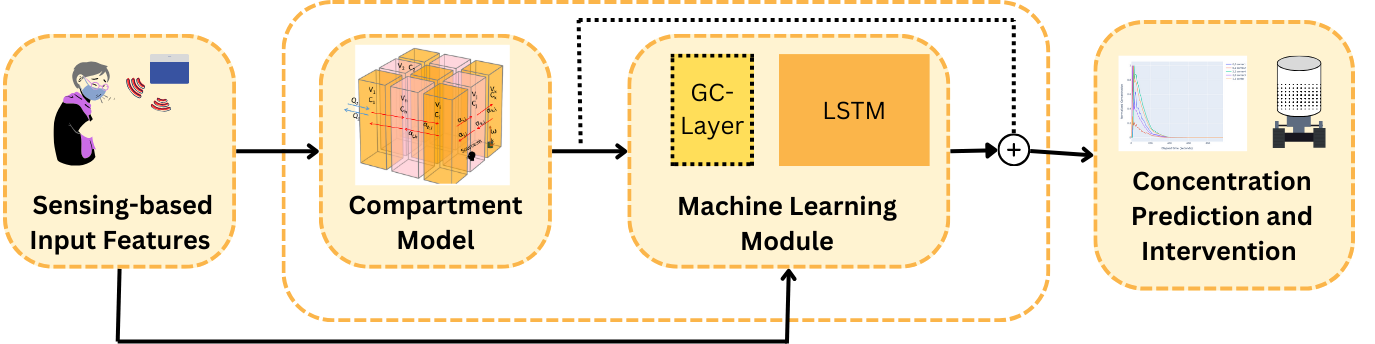}
    \caption{Model schematic using a compartment-LSTM hybrid model with graph convolution layer}
    \label{fig:comp-gc-lstm}
    \vspace{-0.5cm}
\end{figure*}

Our pipeline includes sensing to detect syndromic events, modeling the dispersion of aerosols, and finally mitigating by action from the air-purifier agent.

\subsection{Cough detection model}

We detect cough events as the first step, employing a fine-tuned VGGish model~\cite{VGGish}. The dataset and augmentation technique follow \cite{fluSense}, and our model outperforms the original across all test conditions, including real-world datasets~\cite{fluSense}. Table~\ref{tab:coughModel} shows our model's performance compared to the original model deployed in \cite{fluSense}

\begingroup
\renewcommand{\arraystretch}{0.8}
\begin{table}[!htbp]
\caption{Performance comparison of our VGGish-based cough classifier with the classifier described in~\cite{fluSense} }
\label{tab:coughModel}
    \renewcommand{\arraystretch}{1.1} 
    \resizebox{\columnwidth}{!}{%
    \begin{tabular}{|l|l|l|l|l|l|l|}
        \hline
        \multirow{2}{*}{\textbf{Testing sound type}} & \multicolumn{3}{c|}{\textbf{Current model (VGGish)}} & \multicolumn{3}{c|}{\textbf{ \cite{fluSense}}} \\
        \Xcline{2-7}{.5pt}
        & \textit{\textbf{R (\%)}} & \textit{\textbf{P (\%)}} & \textit{\textbf{F1 (\%)}} & \textit{\textbf{R (\%)}} & \textit{\textbf{P (\%)}} & \textit{\textbf{F1 (\%)}} \\ \hline
     
No Background Noise & 91.5 & 91.5 & 91.5 & 90.2 & 90.2 & 90.2 \\ \hline
With Speech  & 87   &   85.5   &  86 & 82.4 & 82.3 & 82.4 \\ \hline
With Hospital Noise & 87   & 88  & 86 & 84.5 & 85.4 & 84.4 \\ \hline
With All Augmentations & 89.5 & 90.5 & 89.5 & 87 & 87.3 & 86.9\\ \hline
FluSense Dataset \cite{fluSense} & 93.1 & 93.2 & 93 & 89 & 87 & 88\\ \hline
\end{tabular}
   }

\end{table}
\endgroup

\subsection{Residence Time Distribution (RTD) Analysis}
\label{sec:rtd-analysis}

RTD analysis measures the concentration $C(t)$ of a tracer in the air exhaust of the room over time using sensor readings. The time series of the concentration at the outlet is used to compute RTD metrics, such as cumulative RTD, defined in equation (\ref{eq:crtd}).

\begin{equation}
\label{eq:crtd}
    F(t) = \int_{0}^{t} \frac{C(t)}{\int_{0}^{\infty} C(t) dt} dt
\end{equation}

We calculate the mean residence time (MRT) using PM sensor data by measuring the duration for concentrations to revert to initial levels after aerosol-generating events during our experiments. This involves establishing baseline concentrations prior to emulated coughs and tracking the time required for concentrations to return to initial levels.

\subsection{Aerosol Concentration Models}

Data from the testbed is used to predict aerosol concentrations over time. Pre-processing involves normalization, and the core of our model is a physics-based \emph{compartment model}, dividing the space into compartments. The dynamics of each compartment are governed by the following equation \ref{eq:multi-comp} which captures the exchange of aerosol mass (denoted as $C_i$) between compartments, and any aerosol source and exhaust.

\begin{equation}
    V_i\frac{{dC_i}}{{dt}} = \sum_{j=1}^{|N_i|} \left( \alpha_{j,i} \cdot C_j - \alpha_{i,j} \cdot C_i \right) -\gamma \cdot \omega \cdot C_i - Q \cdot C_i + \dot{m}
    \label{eq:multi-comp}
\end{equation}
where:
\begin{description}
    \item[$V_i$]: Volume of compartment $i$
    \item[$C_i$]: PM Concentration in compartment $i$
    \item[$N_i$]: Set of neighbor compartments of $i$ where $C_j \in N_i$
    \item[$\alpha_{j,i}$]: Outflow rate from neighbor compartment $j$ to $i$
    \item[$\dot{m}$]: Source aerosol release rate into compartment $i$
    \item[$Q$]: Rate of exhaust output from compartment $i$
    \item[$\gamma$]: Filter pollutant removal efficiency rate
    \item[$\omega$]: Rate of air going through filter unit in $i$
\end{description}

Our nine-compartment model assumes adjacency without diagonal connections. The rate parameters were learned using collected data without an active air filter employing \emph{differential evolution} \cite{storn1997differential} method since this provided improved results compared with other strategies such as gradient-based methods and dual annealing. We extend the compartment model with machine learning modules, incorporating LSTM and graph convolution to improve aerosol concentration predictions. Inputs include the cough, air purifier, and AC locations; time step encoding, and compartment model output. Directly applying the model requires re-estimating parameters with each new purifier position. We train these LSTM models using the mean squared error (MSE) loss function.

The \emph{Compartment-LSTM hybrid model} integrates the compartment model with an LSTM module. We explore two configurations: directly predicting concentrations (Comp-LSTM), and predicting errors in the compartment model (Comp-LSTM-Res) output through a residual connection. Additionally, the \emph{Compartment-GC-LSTM} model combines LSTM with graph convolutional layers, capturing spatiotemporal relationships more effectively. Fig.~\ref{fig:comp-gc-lstm} visually represents the hybrid models and their components.

\subsubsection{Adaptive Model with Model-Agnostic Meta-Learning}
To adapt our model to diverse conditions, we train our models using first-order model agnostic meta-learning (MAML) \cite{finn2017model}. MAML-based training comprises two phases where in the first phase, the meta-learner's parameters are randomly initialised. The training process involves multiple learning episodes, each representing a different potential scenario (e.g., various room and HVAC configurations, inclusion of air purifier, furniture arrangements, and locations of the coughing agent). We update the weights of the base meta-learner model based on these learning episodes. For each episode, we utilize a small dataset sampled from the full dataset, to represent a distinct task. Each episode's dataset is divided into a support set and a query set. Within each episode, we adapt a copy of the meta-learner by updating parameters to the specific task through an inner loop of gradient updates based on the support set. Subsequently, the meta-learner model parameters are updated based on the performance of the episode-specific adapted models on the query set. This process ensures the meta-learner is optimized for adaptation to new tasks, even with limited data.  During the second phase, the meta-trained model is introduced to a new, unseen task. The model can adapt its parameters to this new task relatively faster through a few gradient steps with limited data samples in the few shot manner, leveraging the adaptability.

\subsection{Optimal Air Purifier Placement Model}
The concentration model drives the purifier's placement strategy to minimize aerosol residence time. The purifying agent autonomously moves within the grid and adjusts fan speed, optimizing power usage and filter life. Cough events and grid-level localization inform purifier positioning, utilizing an environment map that includes the positions of vents, furniture, and obstacles while the optimal placement policy is determined using Algorithm~\ref{alg:filter_agent}.

\begin{algorithm}[!h]
    \small
    \caption{Purifier Agent Action Model}\label{alg:filter_agent}
    \KwData{Cough event detection, aerosol concentrations, ventilation patterns}
    \KwResult{Purifier agent action}
    Define tolerance threshold $\tau$\;
    \If{cough event detected}{
        Estimate event location $(x, y)$ within the grid\;
        \ForEach{accessible location $(x_i, y_i)$}{
            Predict concentration $C_{i}(t)$ for purifying location $(x_i, y_i)$ using digital twin model\;
            Calculate residence time $R_{(x_i, y_i)}(t)$ from $C_{i}(t)$\;  
        }
        Identify candidate locations: $L_{\text{c}} = \{(x_i, y_i) : R_{(x_i, y_i)}(t) \leq \min (R_{(x_i, y_i)}(t)) + \tau\}$\;
        Select location $l = \arg\min(\text{distance}(L_{\text{c}}))$\;
        Move agent to location $l$\;
    }
    \Else{
        \If{PM sensor included with the purifier agent}{
            Turn off filter if $C_i < threshold$\;
        }
        \Else{
            Set filter to lowest setting if time since last cough $> \sum_{k=1}^{n} \text{R}_{k}(t)$\;
        }
    }
\end{algorithm}

\section{RESULTS}

\begin{figure*}[!h]
\vspace{+2mm}
    \centering
    \includegraphics[trim={0, 5cm, 0, 5cm}, clip, width=\linewidth]{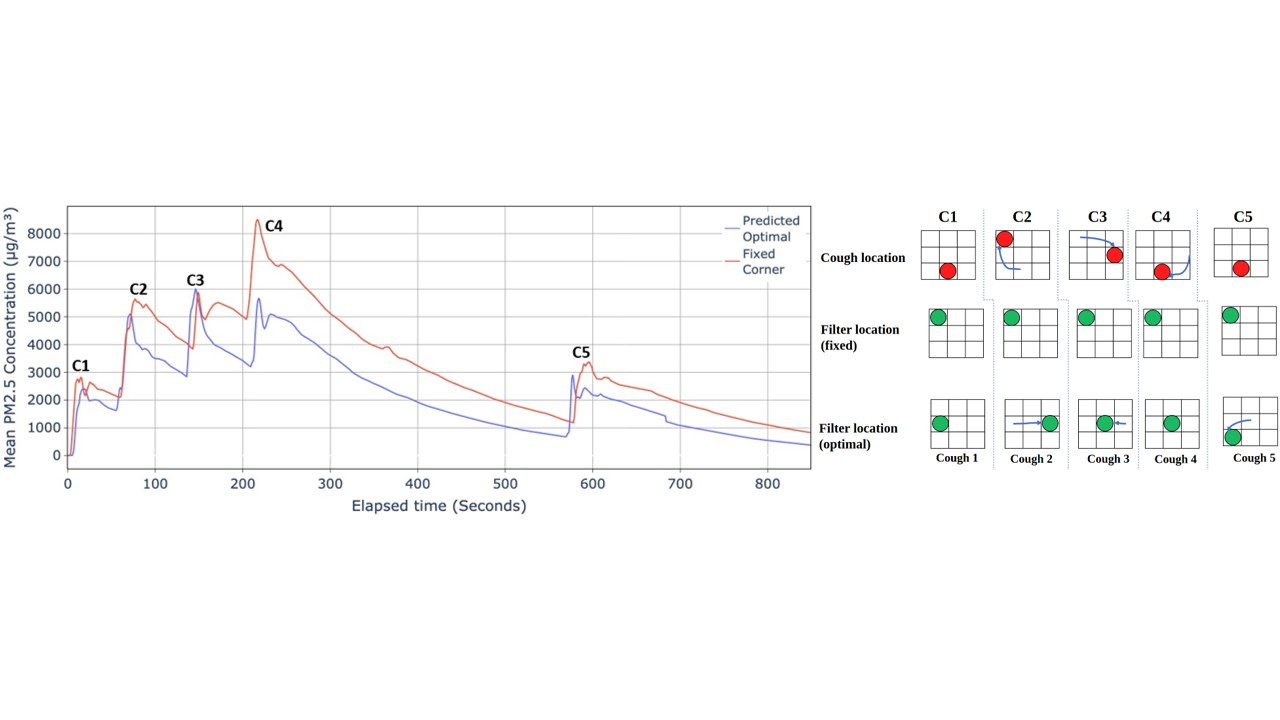}
    \caption{Comparison of PM2.5 concentrations for fixed vs. predicted optimal filter locations. Average PM2.5 levels and filter/cough positions are shown, demonstrating faster concentration reduction with optimal placement.}
    \label{fig:multicough-raw-data-example}
    \vspace{-1em}
\end{figure*}

We evaluated our concentration prediction model using metrics such as mean squared error (MSE), mean absolute error (MAE), and Pearson correlation coefficient ($\rho$). Our primary metric, mean residence time error (MRTE), assesses the system's accuracy in predicting aerosol persistence. The MRTE is calculated as the absolute difference between the predicted and ground-truth residence time (MRT, discussed in Section \ref{sec:rtd-analysis}) across all locations. We use \emph{five-fold cross-validation} and report the average error across all folds.

\subsection{Aerosol concentration model results}

Our baseline models include a support vector regressor (SVR), decision tree regressor (DTR), gradient boosting regressor (GBR), LSTM-baseline model, and a standalone multi-compartment model. We observe improved performance for our hybrid models incorporating the compartment model and machine learning modules compared to the baseline compartment model which uses the rates estimated without air-purifier data. These rates are not re-calibrated since each position of the filter requires separate data collection and optimization. We observe that the predictions from this model are greatly improved by our machine-learning modules. While the other baseline models showed better correlation coefficients than the baseline compartment model (except LSTM), their MRTE performance was notably inferior, as evident in Table~\ref{tab:concentration-pred-result-maml}.

\begingroup
\renewcommand{\arraystretch}{0.75}
\begin{table}[!htbp]
\centering
\caption{Results for baseline models and deep learning-based models (hybrid models trained with MAML approach)}
\label{tab:concentration-pred-result-maml}
\resizebox{\columnwidth}{!}{%
    \begin{tabular}{@{}lcc|ccccc@{}}
    \toprule
    \textbf{Base} & \textbf{ML Module} & \textbf{Residual} & \textbf{MAE} & \textbf{MSE} & $\boldsymbol{\rho}$ & \textbf{MRTE} \\ 
    \midrule
    \textit{Multi-compartment} & - & - & 0.090 & 0.039 & 0.487 & 181.950 \\
    \midrule
    - & \textit{SVR} & - & 0.111 & 0.024 & 0.812  & 353.26 \\
    \midrule
    - & \textit{DTR} & - & 0.078 & 0.019  & 0.820 & 270.55\\
    \midrule
    - & \textit{GBR} & - & 0.071 & 0.019 &  0.831 & 216.249\\
    \midrule
    - & \textit{LSTM} & - & 0.143 & 0.055 & 0.021 & 479.783 \\
    \midrule
    \multirow{2}{*}{\textit{\begin{tabular}[c]{@{}l@{}}Multi-\\ compartment\end{tabular}}} & \multirow{2}{*}{LSTM} & - & 0.066 & 0.017  & 0.843 & 92.325 \\
    & & $\checkmark$ & 0.062 & 0.017 & 0.846 & 79.365  \\
    \midrule
    \multirow{2}{*}{\textit{\begin{tabular}[c]{@{}l@{}}Multi-\\ compartment\end{tabular}}} & \multirow{2}{*}{GC-LSTM} & - & 0.057 & 0.015 & 0.870 & 75.663 \\
    & & $\checkmark$ & 0.061 & 0.016 & 0.850 & 34.191 \\ 
    \bottomrule
    \end{tabular}
}
\vspace{-0.25cm}
\end{table}
\endgroup

The hybrid models, as shown in Table~\ref{tab:concentration-pred-result-maml}, outperform the baseline models across all metrics. The baseline LSTM model performs the worst on our dataset. Models that directly predict the concentration have a better correlation with ground truth, however, models that predict the errors perform better at predicting mean residence time. The learning episodes for meta-training scenarios involving the purifier are segmented based on the row in the grid where the air purifier is located. Our experimental results demonstrate improved results for this approach compared to models trained using standard training without this strategy.

\begingroup
\renewcommand{\arraystretch}{0.75}
\begin{table}[]
\vspace{2mm}
\centering
\caption{Results adapting to different settings using the MAML-trained model}
\label{tab:fewshot-maml-results}
\resizebox{\columnwidth}{!}{%
    \begin{tabular}{@{}ccccccc@{}}
    \toprule
    \textbf{Model}                & \textbf{Change in Setup}      & \textbf{k-shot} & \textbf{MAE} & \textbf{MSE} & $\boldsymbol{\rho}$ & \textbf{MRTE} \\ \midrule
    \multirow{7}{*}{Comp-GC-LSTM} & -                             & -               & 0.061        & 0.016        & 0.85         & 34.191        \\ \cmidrule(l){2-7}
                                  & \multirow{2}{*}{(+)Furniture} & 0               & 0.122        & 0.043        & 0.822        & 106.07        \\
                                  &                               & 2               & 0.092        & 0.038        & 0.834        & 64.34         \\ \cmidrule(l){2-7}
                                  & \multirow{2}{*}{AC location}  & 0               & 0.211        & 0.054        & 0.786        & 208.16        \\
                                  &                               & 2               & 0.144        & 0.044        & 0.811        & 135.232       \\ \cmidrule(l){2-7}
                                  & \multirow{2}{*}{AC fan speed} & 0               & 0.31         & 0.134        & 0.664        & 131.776       \\
                                  &                               & 2               & 0.092        & 0.041        & 0.822        & 72.84         \\ \bottomrule
    \end{tabular}
}
\vspace{-0.75cm}
\end{table}
\endgroup

\textbf{Results in Zero-Shot and Few-Shot Settings}: To ensure real-world adaptability, we evaluated our models using MAML-based training under modified testbed conditions. Table~\ref{tab:fewshot-maml-results} presents results across three variations: added furniture (two chairs, a large cardboard box, and a file cabinet alongside existing chair and garbage bin), altered AC location, and varying fan speeds. Objects were repositioned across trials, and results reflect mean performance over multiple setups. Despite the environmental changes, our model demonstrated robust zero-shot and few-shot performance without significant degradation compared to baselines. With just two examples, it adapted effectively, highlighting strong generalization with limited fine-tuning. Consistent trends were observed across our other models - for example, the Comp-LSTM model exhibited MRTE reductions of 28\% for furniture addition and 38\% for AC location change under 2-shot adaptation (We omit these supplementary results due to space constraints).


\subsection{Optimal Action prediction for Purifier agent}
\begin{figure}[!htbp]
\vspace{1.5mm}
    \centering
    \begin{subfigure}{0.55\linewidth}
        \includegraphics[width=\linewidth]{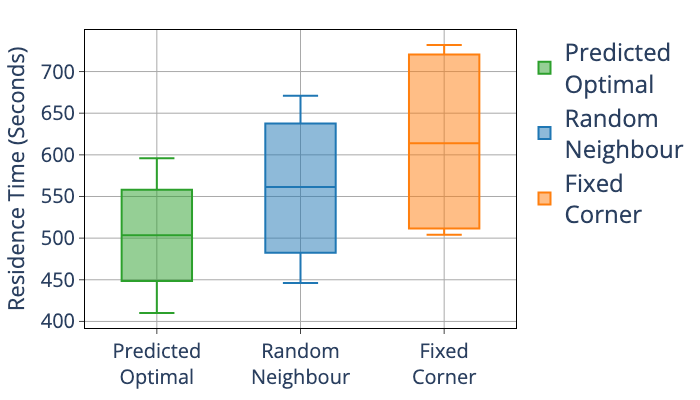}
    \caption{}
    \end{subfigure}
    \begin{subfigure}{0.43\linewidth}
        \includegraphics[width=\linewidth]{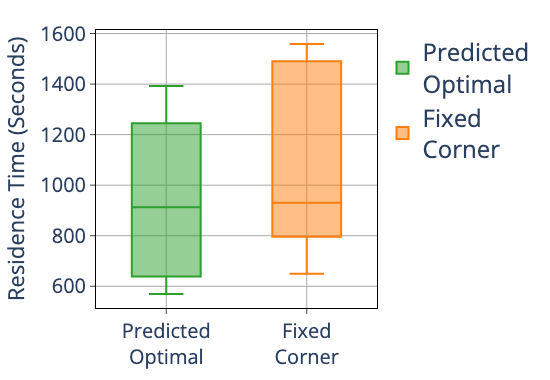}
        \caption{}
    \end{subfigure}
    \caption{Residence time for purifier placement strategies: (a) single, (b) multiple coughs. The optimal strategy uses Algorithm~\ref{alg:filter_agent}; the random neighbor strategy selects a neighboring cell, and fixed position refers to corner placement.}
    
    \label{fig:control-res-time-single cough}
\vspace{-0.5cm}
\end{figure}

We compare our predicted placement strategy for the air purifier robot with other strategies such as placement on adjacent locations of cough, and fixed corners based on electrical outlets. The results from our experiment are displayed in Fig.~\ref{fig:control-res-time-single cough}. The fixed corner placement performs worst in terms of mean residence time, followed by the random neighbor strategy. This demonstrates that placing the filter near the aerosol source is effective, however, this is not always the case (refer to Section \ref{sec:Simulation}). We observe a similar pattern in the multi-cough setup.

Fig.~\ref{fig:multicough-raw-data-example} demonstrates a comparison of our predicted optimal strategy with the other ones in our testbed environment. We repeated the cough emulation at the same time across two trials for different filter placements. We observe that the average concentration increases similarly in both cases. However, the deployment of the purifier in locations recommended by our system can lower the concentration faster, resulting in an overall lower mean residence time.

\section{DISCUSSION}

\subsection{Validation of Cough Generation}
Our cough generation technique is consistent with established methodologies in the literature \cite{zhou_development_2022}. Our robotic mannequin replicates human cough properties with high fidelity, producing cough events lasting 0.9–1.0 seconds, closely matching the state-of-the-art cough simulators, which generate cough between 0.7–0.9 seconds. Key parameters such as a cough flow rate of 2.6 L/s, cough volume ranging from 1.8 to 2.4 L, a mouth size of 3.8 $cm^2$, and a horizontal cough distance of 2.5 m further align with the literature. Additionally, particle sizes produced by our emulator, categorized into bins of 1.0, 2.5, 4.0, and 10.0 microns, show close correspondence with data from other studies \cite{cough_size}. 

\subsection{Software Simulation of Cough Plumes}
\label{sec:Simulation}
\begin{figure}[!h]
    \centering
    \includegraphics[trim={16cm 3cm 14cm 7cm}, clip, width=0.8\linewidth]{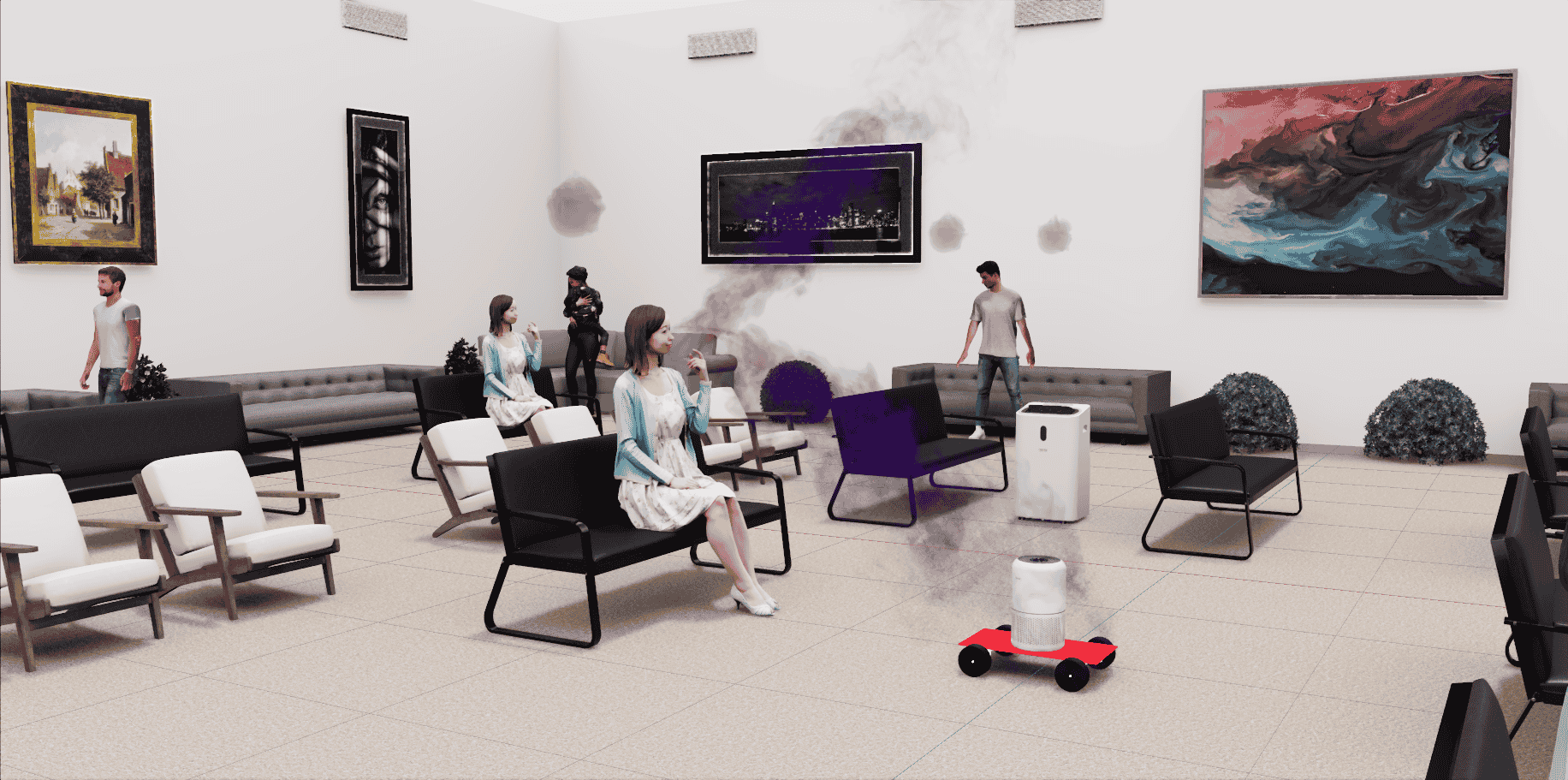}
    \caption{Complex trajectory of the aerosol, influenced by airflow characteristics within the environment, revealing zones with lingering aerosol, demonstrating potential sub-optimal residence time despite the purifier's placement.}
    \label{fig:Omniverse_Simulations}
    \vspace{-0.3cm}
\end{figure}

Leveraging NVIDIA Omniverse\cite{omniverse-simulation}, we performed high-fidelity simulations of cough plumes in a custom-built hospital waiting room environment, featuring complex air outlets and purifier configurations, as observed in Fig.\ref{fig:Omniverse_Simulations}. Employing techniques from \cite{gpu_gems, gpu_gems_3} along with the Omniverse Flow extension, our simulations comprehensively replicate aerosol particle dynamics by incorporating velocity and pressure gradients, vortex formations, and buoyant forces. Our qualitative analysis reveals that although positioning the air purifier directly in the cough's path captures a substantial fraction of the aerosol, it does not fully mitigate dispersion due to ambient air currents and ventilation effects. The simulation demonstrates that aerosol plumes, influenced by ceiling vents and HVAC systems, disperse in complex, multi-directional patterns, challenging the assumption that direct interception is optimal. These results highlight the necessity for advanced strategies in purifier placement and ventilation to effectively minimize aerosol residence time in dynamically ventilated environments. Future work will focus on optimizing these strategies to enhance real-world application efficacy.

\begin{figure}[!h]
    \centering
    \begin{subfigure}{0.35\linewidth}
        \includegraphics[width=\linewidth, height=1.2in]{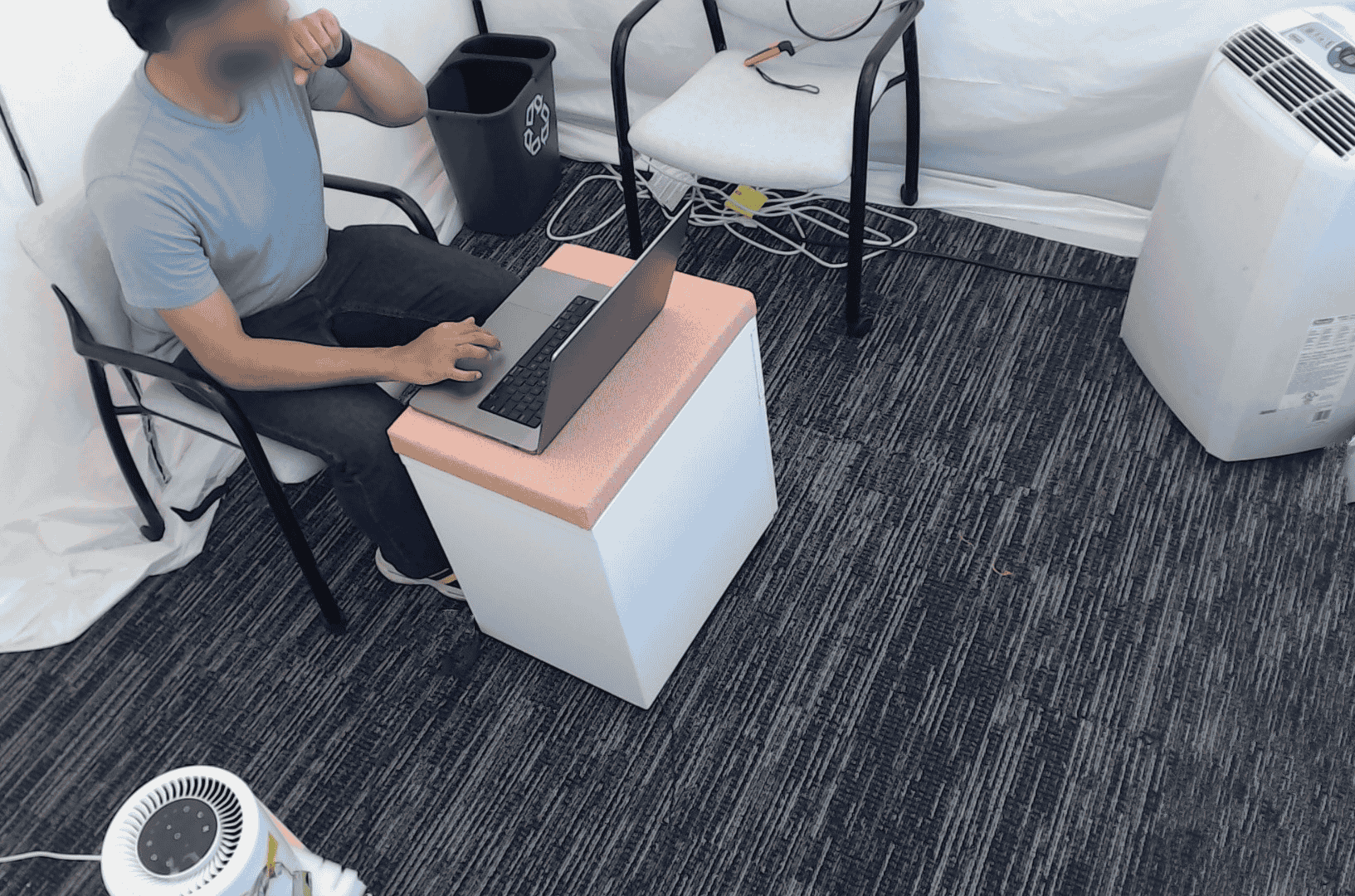}
    \caption{}
    \end{subfigure}
    \begin{subfigure}{0.59\linewidth}
        \includegraphics[width=\linewidth, height=1.2in]{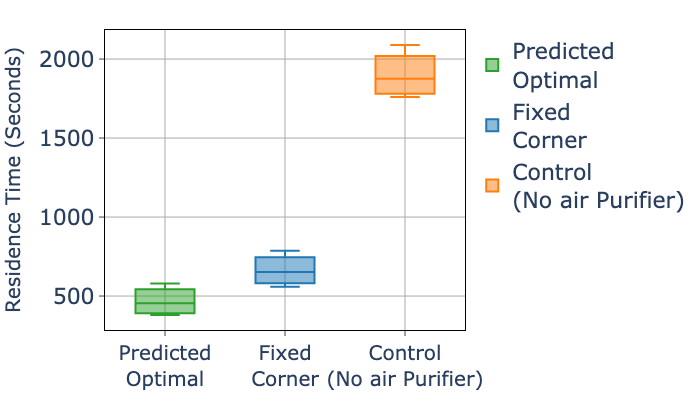}
        \caption{}
    \end{subfigure}
    \caption{Results from our deployment in a human-occupied space (A fraction of our testbed): (a) Sample image from a session; (b) Prolonged aerosol presence without filtration (control) vs. reduced residence time with optimal placement.}
    \label{fig:human-exp}
    \vspace{-0.6cm}
\end{figure}%
\subsection{Deployment in Human-Occupied Space}
\label{sec:human-study-description}

An ongoing IRB-approved study to evaluate our system is under progress; the nascent findings with a single consenting participant are presented here. In an office-like environment, the participant was asked to cough multiple times during sessions lasting approximately one hour. We then analyzed residence time with and without intervention. We observe consistent results from our real-world deployment as shown in Fig. ~\ref{fig:human-exp}. We observe a lower number in terms of the peak concentration compared with the coughing robot since we have a healthy participant mimicking cough action. The data indicate that concentrations remain elevated for a longer period in the control case (no purifier intervention). Using the predicted optimal strategy for the air filter facilitates quicker elimination of emitted particles compared to fixed corner placement.

\section{Acknowledgments}
This work was supported by grants from the Halıcıoğlu Data Science Institute at the University of California, San Diego.

\addtolength{\textheight}{-7.62cm}   
\bibliographystyle{IEEEtran}
\bibliography{IEEEabrv, AeroSafe_Ref}

\end{document}